\documentclass[11pt,a4paper]{article}
\usepackage[hyperref]{naaclhlt2019}
\usepackage{times}
\usepackage{latexsym}
\usepackage{graphicx}
\usepackage{amsfonts}
\usepackage{amsmath}
\usepackage{wasysym}
\usepackage{hyperref}
\usepackage{verbatim}

\usepackage{booktabs}
\usepackage{multirow}

\usepackage{url}

\aclfinalcopy 

\title{Probing Biomedical Embeddings from Language Models}
\author{Qiao Jin \\
  University of Pittsburgh \\
  {\tt qiao.jin@pitt.edu} \\\And
  Bhuwan Dhingra \\
  Carnegie Mellon University \\
  {\tt bdhingra@cs.cmu.edu} \\\AND
  William W. Cohen \\
  Google, Inc. \\
  {\tt wcohen@google.com} \\\And
  Xinghua Lu \\
  University of Pittsburgh \\
  {\tt xinghua@pitt.edu} \\
  }

\date{}

\begin{document}

\setlength{\abovedisplayskip}{4pt}
\setlength{\belowdisplayskip}{4pt}
\setlength\belowcaptionskip{-3ex}

\maketitle
\begin{abstract}
Contextualized word embeddings derived from pre-trained language models (LMs) show significant improvements on downstream NLP tasks.
Pre-training on domain-specific corpora, such as biomedical articles, further improves their performance.
In this paper, we conduct probing experiments to determine what additional information is carried \textit{intrinsically} by the in-domain trained contextualized embeddings.
For this we use the pre-trained LMs as fixed feature extractors and
restrict the downstream task models to not have additional sequence modeling layers.
We compare BERT \cite{devlin2018bert}, ELMo \cite{peters2018deep}, BioBERT \cite{lee2019biobert} and BioELMo, a biomedical version of ELMo trained on $10$M PubMed abstracts.
Surprisingly, while fine-tuned BioBERT is better than BioELMo in biomedical NER and NLI tasks, as a fixed feature extractor BioELMo outperforms BioBERT in our probing tasks. We use visualization and nearest neighbor analysis to show that better encoding of entity-type and relational information leads to this superiority.
\end{abstract}

\section{Introduction}
NLP has seen an upheaval in the last year, with contextual word embeddings, such as ELMo \citep{peters2018deep} and BERT \citep{devlin2018bert}, setting state-of-the-art performance on many tasks.
These empirical successes suggest that unsupervised pre-training from large corpora could be a vital part of NLP models. 
In specific domains like biomedicine, NLP datasets are much smaller than their general-domain counterparts\footnote{For example, MedNLI \cite{romanov2018lessons} only has about $11$k training instances while the general domain NLI dataset SNLI \cite{bowman2015large} has $550$k.}, which leads to a lot of ad-hoc models: some infer through knowledge bases \citep{chandu2017tackling}, while others leverage large-scale general domain datasets for domain adaptation \cite{wiese2017neural}. However, unlabeled biomedical texts are abundant, and their full potential has perhaps not yet been fully realized.

We train a domain-specific version of ELMo on $10$M PubMed abstracts, called BioELMo\footnote{Available at \url{https://github.com/Andy-jqa/bioelmo}.}.
Experiments on biomedical named entity recognition (NER) dataset BC2GM \cite{smith2008overview} and biomedical natural language inference (NLI) dataset MedNLI \cite{romanov2018lessons} clearly show the utility in training in-domain contextual word representations, but we would also like to know exactly what extra information is carried \textit{intrinsically} in these embeddings.

To answer this question, we design two \textit{probing tasks}, one for NER and one for NLI, where contextualized embeddings are used solely as fixed feature extractors and no sequence modeling layers are allowed above the embeddings. This setting prohibits the model from capturing task-specific contextual patterns, and instead only utilizes the information already present in the representations. In parallel to our work of BioELMo, \citet{lee2019biobert} introduce BioBERT, which is a biomedical version of in-domain trained BERT. We also probe BioBERT in our experiments.


Expectedly, BioELMo and BioBERT perform significantly better than their general-domain counterparts.
When fine-tuned, BioBERT outperforms BioELMo, however, when used as fixed feature extractors, BioELMo is better than BioBERT in our probing tasks. Visualizations and nearest neighbor analyses suggest that it's because BioELMo more effectively encodes entity-types and information about biomedical relations, such as disease and symptom interactions, than BioBERT.



\section{Related Work}

\paragraph{Embeddings from Language Models:} ELMo \cite{peters2018deep} is a pre-trained deep bidirectional LSTM (biLSTM) language model. ELMo word embeddings are computed by taking a weighted sum of the hidden states from each layer of the LSTM. The weights are learned along with the parameters of a task-specific downstream model, but the LSTM layers are kept fixed.
Recently, \citet{devlin2018bert} introduced BERT, and they showed that pre-training transformer networks on a masked language modeling objective leads to even better performance by fine-tuning the transformer weights on a broad range of NLP tasks. We study biomedical in-domain versions of these contextualized word embeddings in comparison to the general ones.

\paragraph{Biomedical Word Embeddings:} Context-independent word embeddings, such as word2vec (w2v) \citep{mikolov2013distributed} trained on biomedical corpora, are widely used in biomedical NLP models. Some recent works reported better NER performance with in-domain trained ELMo than general ELMo \cite{zhu2018clinical,sheikhshabbafghi2018domain}. \citet{lee2019biobert} introduce BioBERT, which is BERT pre-trained on biomedical texts and set new state-of-the-art performance on several biomedical NLP tasks. We reaffirm these results on biomedical NER and NLI datasets with in-domain trained contextualized embeddings, and further explore \textit{why} they are superior.

\paragraph{Probing Tasks:} Designing tasks to probe sentence or token representations for linguistic properties has been a widespread practice in NLP. InferSent \cite{conneau2017supervised} uses transfer tasks to probe for sentence embeddings pre-trained on supervised data. Many studies \cite{dasgupta2018evaluating,poliak2018collecting} design new test sets to probe for specific linguistic signals in sentence rerpesentations. Tasks to probe for token-level properties are explored by \citet{blevins2018deep,peters2018dissecting}, where they test whether token embeddings from different pre-training schemes encode part-of-speech and constituent structure. 

\citet{tenney2018you} extend token-level probing to span-level probing and consider a broader range of tasks. Our work is different from them in the following ways -- (1) We probe for biomedical domain-specific contextualized embeddings and compare them to the general-domain embeddings; (2) For NER, instead of classifying the tag for a given span, we adopt an end-to-end setting where the spans must also be identified. This allows us to compare the probing results to state-of-the-art numbers; (3) We also probe for relational information using the NLI task in an end-to-end style.
 
\section{Methods}
\subsection{Biomedical Contextual Embeddings}
\paragraph{BioELMo:} We train BioELMo on the \textbf{PubMed} corpus.
PubMed provides access to MEDLINE, a database containing more than $24$M biomedical citations\footnote{\url{https://www.ncbi.nlm.nih.gov/pubmed/}}. We used $10$M recent abstracts ($2.46$B tokens) from PubMed to train BioELMo. The statistics of this corpus are very different from more general domains.
For example, the token \textbf{patients} ranks 22 by frequency in the PubMed corpus while it ranks 824 in the 1B Word Benchmark dataset \cite{chelba2013one}.
We use the Tensorflow implementation\footnote{\url{https://github.com/allenai/bilm-tf}} of ELMo to train BioELMo. We keep the default hyperparameters and it takes about $1.7$K GPU hours to train 8 epochs. BioELMo achieves an averaged forward and backward perplexity of 31.37 on test set.

\paragraph{BioBERT:} In parallel to our work, \citet{lee2019biobert} developed BioBERT, which is pre-trained on English Wikipedia, BooksCorpus and fine-tuned on PubMed ($7.8$B tokens in total). BioBERT was initialized with BERT and further trained on PubMed for $200$K steps.%
\footnote{We note there is a difference in the size of training corpora for BioBERT and BioELMo, but since we trained BioELMo before BioBERT
was available, we could not control for this difference.}

To get fixed features of tokens, we use the learnt downstream task-specific layer weights to calculate the average of 3 layers (1 token embedding layer and 2 biLSTM layers) for BioELMo and 13 layers (1 token embedding layer and 12 transformer layers) for BioBERT. As fixed feature extractors, BioELMo and BioBERT are not fine-tuned by downstream tasks.

\subsection{Downstream Tasks} \label{whole}

We first use BioELMo with state-of-the-art models and fine-tune BioBERT on the downstream tasks, to test their full capacity.
In \S \ref{probing} we introduce our probing setup which tests BioBERT and BioELMo as fixed feature extractors.

\textbf{NER:} For BioELMo, following \citet{lample2016neural}, we use the contextualized embeddings and a character-based CNN for word representations, which are fed to a biLSTM, followed by a conditional random field (CRF) \cite{lafferty2001conditional} layer for tagging. For BioBERT, we use the single sentence tagging setting described in \citet{devlin2018bert}, where the final hidden states of each token are trained to classify its NER label.

\textbf{NLI:} For BioELMo, We use the ESIM model \cite{chen2016enhanced}, which encodes the premise and hypothesis using biLSTM. The encodings are fed to a local inference layer with attention, another biLSTM layer and a pooling layer followed by softmax for classification. For BioBERT, we use the sentence pair classification setting described in \citet{devlin2018bert}, where the final hidden states of the first token (special `[CLS]') are trained to classify the NLI label for the sentence pair.

\subsection{Probing Tasks} \label{probing}
We design two probing tasks where the contextualized embeddings are only used as fixed feature extractors and restrict the down-stream models to be non-contextual, to investigate the information intrinsically carried by them.
One task is on NER to probe for entity-type information, and the other is on NLI to probe for relational information.

\begin{figure*}
    \centering
    \includegraphics[width=0.75\textwidth]{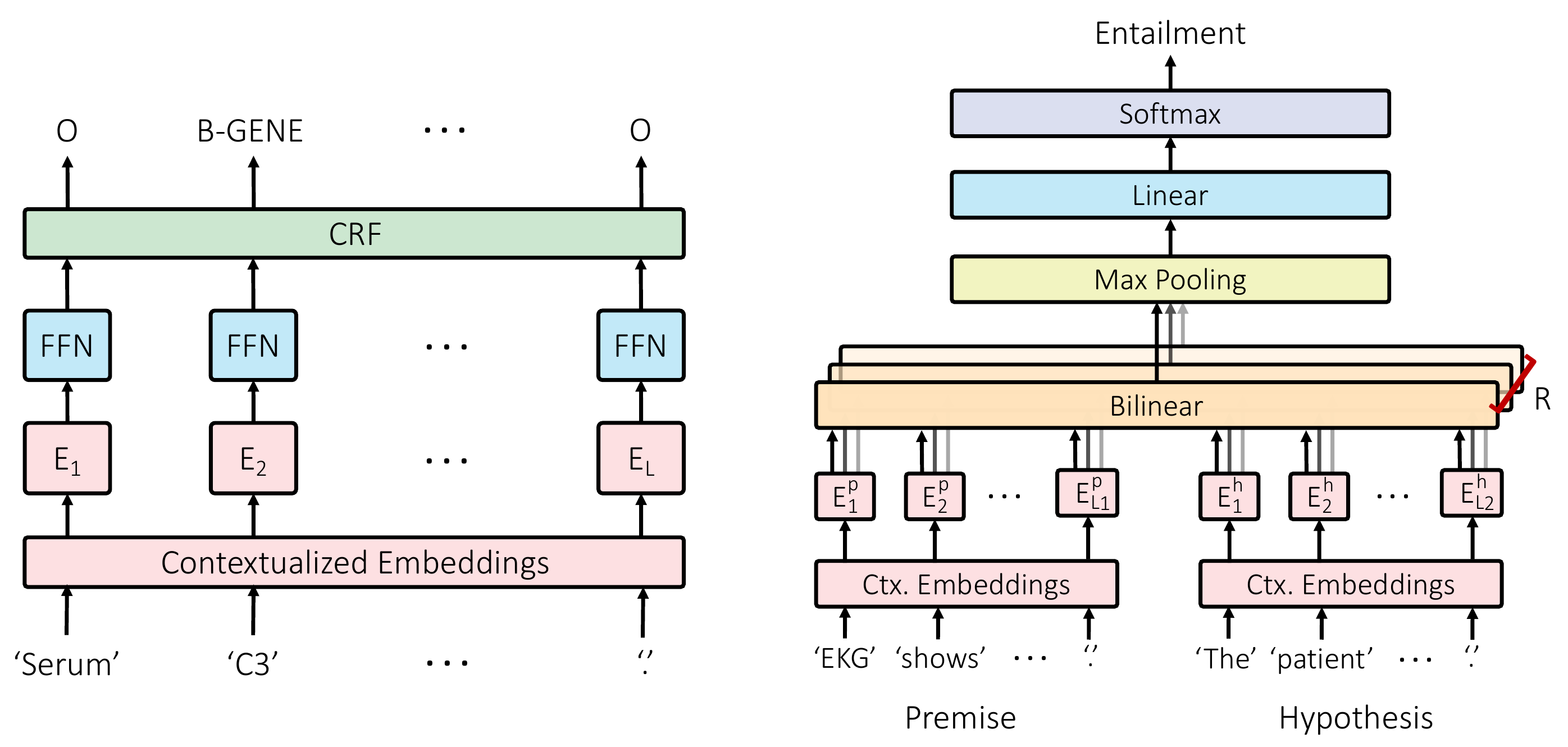}
    \caption{\textbf{Left:} NER probing task. The contextual word representations are directly used to predict the NER labels, followed by a CRF layer to ensure label consistency.
    \textbf{Right:} NLI probing task. Bilinear operators map pairs of word representations to relation representations which are used to predict the entailment label.}
    \label{fig:ner_probe}
\end{figure*}

\textbf{NER Probing Task:} As shown in Figure \ref{fig:ner_probe} (left), 
we embed the input tokens to
$\mathbf{R}=[\mathbf{E_1};\mathbf{E_2};...;\mathbf{E_L}] \in \mathbb{R}^{L \times D_e}$, where $L$ is the sequence length and $D_e$ is embedding size. The embeddings are fed to several feed-forward layers:
\[
\widetilde{\mathbf{E_i}} = \text{FFN}(\mathbf{E_i}) \in \mathbb{R} ^ {T}
\]
where $T$ is the number of tags. $[\widetilde{\mathbf{E_1}};\widetilde{\mathbf{E_2}};...;\widetilde{\mathbf{E_L}}]$ is then fed to a CRF output layer. CRF doesn't model the context but ensures the global consistency across the assigned labels, so it's compatible with our probing task setting.

\begin{figure}
    \centering
    \includegraphics[width=\linewidth]{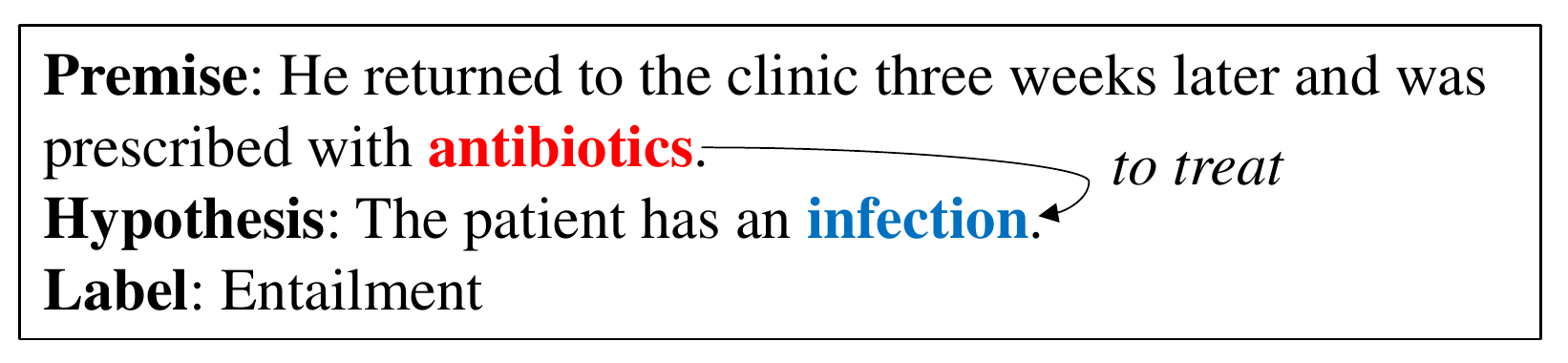}
    \caption{Relation information in a MedNLI instance.}
    \label{fig:mednli_relation}
\end{figure}

\textbf{NLI Probing Task:} Relational information between tokens of premises and hypotheses is vital to solve MedNLI task: as shown in Figure \ref{fig:mednli_relation}, the hypothesis is an entailment because \textbf{antibiotics} are used to treat an \textbf{infection}, which is a drug-disease relation. We design the task shown in Figure \ref{fig:ner_probe} (right) to probe such relational information: We embed the premise and hypothesis seperately to $\mathbf{P} \in \mathbb{R}^{L_1 \times D_e}$ and $\mathbf{H} \in \mathbb{R}^{L_2 \times D_e}$, where $L_1$, $L_2$ are sequence lengths. Then we use bilinear layers\footnote{We also tried models without bilinear layers, which turn out to be suboptimal.} to get $\mathbf{S}=[\mathbf{S_1};\mathbf{S_2};...;\mathbf{S_R}] \in \mathbb{R}^{R \times L_1 \times L_2}$ where
\[
\mathbf{S_r} = \mathbf{P}\mathbf{W_r}\mathbf{H^T} \in \mathbb{R}^{L_1 \times L_2},
\]
and $\mathbf{W_r} \in \mathbb{R}^{D_e \times D_e}$ is the weight matrix of a bilinear layer.
Note that each element of $\mathbf{S_r}$ encodes the interaction between a token from the premise and a token from the hypothesis.
We denote
\begin{equation} \label{eq:1}
\mathbf{h_{ij}} = \begin{bmatrix}\mathbf{S_1}[i,j]&...&\mathbf{S_R}[i,j]\end{bmatrix} ^ T \in \mathbb{R}^{R},
\end{equation}
as the \textbf{distributed relation representation} between token $i$ in premise and token $j$ in hypothesis, and $R$ is the tunable dimension of it. We then apply an element-wise maximum pooling layer:
\[
\widetilde{\mathbf{h}} = \max_{i,j} \mathbf{h_{ij}} \in \mathbb{R}^{R}.
\]
We use a linear layer to compute the softmax logits of the NLI labels, e.g.
$p(\text{entailment}) \propto \text{exp}(\widetilde{\mathbf{h}}^T\,\mathbf{w_{ent}})
$,
where $\mathbf{w_{ent}}$ is the learnt weight vector corresponding to the entailment label.

For BERT, we probe two variants.
The first, denoted as BERT / BioBERT, feeds the premise and hypothesis to the model separately.
The second, denoted as BERT-tog / BioBERT-tog, concatenates the two sentences by the `[SEP]' token and feeds to the model \textit{together} to get the embeddings. This is how BERT is supposed to be used for sentence pair classification tasks, but it's not comparable to ELMo in our setting since ELMo doesn't take two sentences together as input.

\section{Experiments}
\subsection{Experimental Setup}

\noindent \textbf{Data:} For the NER task, we use the BC2GM dataset. BC2GM stands for BioCreative II gene mention dataset \cite{smith2008overview}. The task is to detect gene names in sentences. It contains $15$k training and $5$k test sentences. We also test on the general-domain CoNLL 2003 NER dataset \citep{tjong2003introduction}, where the task is to detect entities such as person and location. 

For the NLI task, we use the MedNLI dataset \cite{romanov2018lessons}, where the task is, given a pair of sentences (premise and hypothesis), to predict whether the relation of entailment, contradiction, or neutral (no relation) holds between them. The premises are sampled from doctors' notes in the clinical dataset MIMIC-III \cite{johnson2016mimic}. The hypotheses and annotations are generated by clinicians. It contains 11,232 training, 1,395 development and 1,422 test instances. We also test on the general-domain SNLI dataset \cite{bowman2015large}, where the premises and hypotheses are drawn from image captions.

\noindent \textbf{Compared Settings:} For each dataset, the \textbf{Whole} setting refers to the state-of-the-art model we used (described in \S\ref{whole}), including contextual modeling layers or fine-tuning of the embedding encoder. \textbf{Probing} and \textbf{Control} settings describe the probing task model introduced in \S\ref{probing}. The control setting tests the representations on a general-domain dataset/task, to check whether we lose any information in domain-specific embeddings. Probing and control results are averaged over three seeds.

\noindent \textbf{Compared Embeddings:} We compare: (1) non-contextual biomedical w2v trained on a biomedical corpus of $5.5$B tokens \cite{moen2013distributional}, (2) ELMo trained on a general-domain corpus of $5.5$B tokens\footnote{\url{https://allennlp.org/elmo}}, (3) BioELMo\footnote{Though BioELMo uses the smallest corpus to train, it performs better than BioBERT in probing setting, and general ELMo in whole and probing setting.}, (4) Cased base version of BERT trained on a general-domain corpus of $3.3$B tokens\footnote{\url{https://github.com/google-research/bert}} and (5) BioBERT\footnote{\url{https://github.com/dmis-lab/biobert}}.

\subsection{Main Results}

\begin{table}[htbp]
\centering
\small
\begin{tabular}{lccc}
\toprule
\multirow{2}{*}{\textbf{Method}} & \multicolumn{3}{c}{\textbf{F1 (\%)}} \\
\cmidrule(l){2-4} 
& \textbf{Whole} & \textbf{Probe} & \textbf{Ctrl.} \\
\midrule
\citet{ando2007biocreative} & 87.2 & -- & --\\
\citet{rei2016attending} & 88.0 & -- & --\\
\citet{sheikhshabbafghi2018domain} & 89.7 & -- & --\\
\midrule
Biomed w2v & 84.9 & 78.5 & 67.5\\
General ELMo & 87.0 & 82.9 & \textbf{84.0}\\
General BERT & 89.2 & 84.9 & 83.6 \\
BioELMo & 90.3 & \textbf{88.4} & 80.9\\
BioBERT & \textbf{90.6} & 88.2 & 83.4\\
\bottomrule
\end{tabular}
\caption{NER test results. \textbf{Whole}: whole model performance on BC2GM; \textbf{Probe}: Probing task performance on BC2GM; \textbf{Ctrl.}: Probing task performance on CoNLL 2003 NER. We use the official evaluation codes to calculate the F1 scores where there are multiple ground-truth tags, so the F1 scores are much higher than what were reported in \citet{lee2019biobert}.}
\label{tab:bc2gm}
\end{table}

\subsubsection{NER Results}
\noindent \textbf{In Domain v.s. General Domain:} Results in Table \ref{tab:bc2gm} show that BioBERT and BioELMo in the Whole setting perform better than the general BERT and ELMo and biomed w2v, setting new state-of-the-art performance for this dataset. 

BioBERT and BioELMo remains competitive in the Probing setting, doing much better than their general domain counterparts and even general ELMo in the Whole setting. This shows that with the right pre-training, the downstream model can be considerably simplified.

Unsurprisingly, in the Control setting BioBERT and BioELMo do worse than their general counterparts, indicating that the gains come at the cost of losing some general-domain information. 
However, the performance gaps (absolute differences) between ELMo and BioELMo are larger in the biomedical domain than it is in the general domain, which is also true for BERT and BioBERT. For ELMo and BioELMo, we believe it is because the PubMed corpus contains many mentions of general-domain entities whereas the reverse is not true. Because BioBERT is initialzied with BERT and also uses general-domain corpora like Enligsh WikiPedia for pre-training, it's not surprising that BioBERT is just $0.2$ worse than BERT on CoNLL 2003 NER in control setting.

\noindent \textbf{BioELMo v.s. BioBERT:} Fine-tuned BioBERT outperforms BioELMo with biLSTM and CRF on BC2GM. As a feature extractor, BioBERT is slightly worse than BioELMo in probing task of BC2GM, but outperforms BioELMo in probing task of CoNLL 2003, which can be explained by the fact that BioBERT is also pre-trained on general-domain corpora.




\begin{table}[htbp]
\centering
\small
\begin{tabular}{lccc}
\toprule
\multirow{2}{*}{\textbf{Method}} & \multicolumn{3}{c}{\textbf{Accuracy (\%)}} \\
\cmidrule(l){2-4} 
& \textbf{Whole} & \textbf{Probe} & \textbf{Ctrl.}\\
\midrule
\citet{romanov2018lessons} & 76.6 & -- & --\\
\midrule
Biomed w2v & 74.2 & 71.1 & 59.2\\
General ELMo & 75.8 & 69.6 & 60.8\\
General BERT & -- & 67.6 & 62.1\\
General BERT-tog & 77.8 & 71.0 & \textbf{74.1}\\
BioELMo & 78.2 & \textbf{75.5} & 58.3\\
BioBERT & -- & 70.1 & 58.8\\
BioBERT-tog & \textbf{81.7} & 73.8 & 69.9\\
\bottomrule
\end{tabular}
\caption{NLI test results. \textbf{Whole}: whole model performance on MedNLI; \textbf{Probe}: Probing task performance on MedNLI; \textbf{Ctrl.}: Probing task performance on SNLI. To make the results comparable, we only use the same number of SNLI training instances as that of MedNLI.}
\label{tab:mednli}
\end{table}

\subsubsection{NLI Results}
\noindent \textbf{In Domain v.s. General Domain:} Table \ref{tab:mednli} shows that BioBERT and BioELMo in the Whole setting perform better than their general domain counterparts and biomedical w2v for NLI, setting state-of-the-art performance for this dataset as well. 

Once again, we observe that BioBERT and BioELMo outperform their general domain counterparts in the Probing settings, which comes at the cost of losing general domain information as indicated in the Control setting results.


Note that the Probing task \textit{only} models relationships between tokens, but we still see competitive accuracy in that setting ($75.5\%$ vs $76.6\%$ previous best).
This suggests that, (i) many instances in MedNLI can be solved by identifying token-level relationships between the premise and the hypothesis, and (ii) BioELMo already captures this kind of information in its embeddings. 


\noindent \textbf{BioELMo v.s. BioBERT:} Fine-tuned BioBERT does much better than BioELMo with ESIM model. However, BioELMo performs better than BioBERT by a large margin in the probing task of MedNLI. We explore this in more detail in the next section. Again, BioBERT is better than BioELMo in probing task of SNLI because it's also pretrained on general corpora.

We notice that the \textit{-tog} setting improves the BERT performance. 
Encoding two sentences separately, BioELMo still outperforms BioBERT-tog. It suggests that BioELMo is a better feature extractor than BioBERT,
even though the latter has superior performance when fine-tuned on MedNLI.

\subsection{Analysis}

\begin{figure*}
    \centering
    \includegraphics[width=\textwidth]{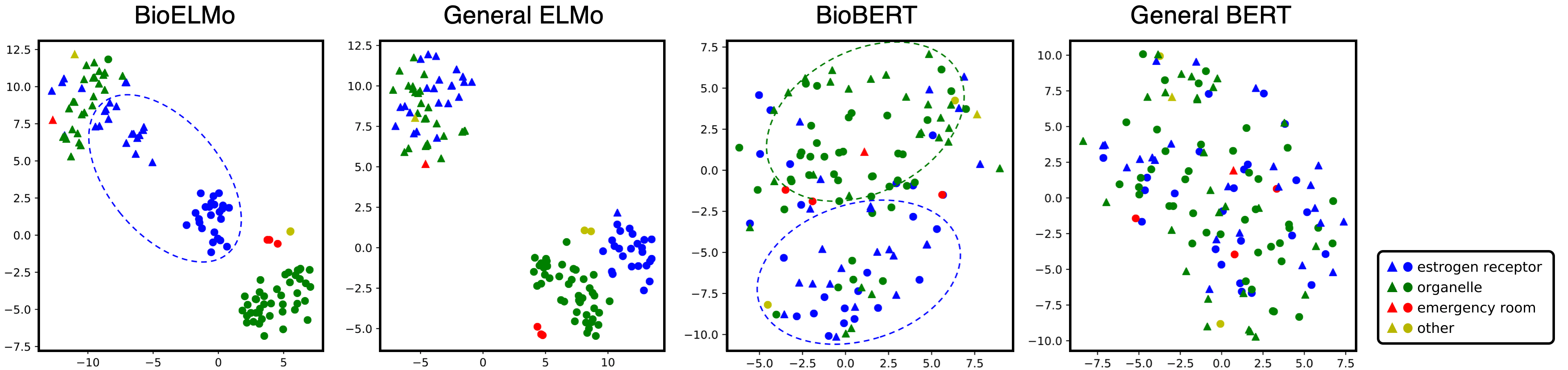}
    \caption{t-SNE visualizations of the token \textbf{ER} embeddings in different contexts by BioELMo, general ELMo, BioBERT and general BERT. \CIRCLE \,and \UParrow \,represent \textbf{ER} mentions within and outside of parentheses, respectively. Colors refer to different actual meanings of the \textbf{ER} mention.}
    \label{fig:entity}
\end{figure*}

\subsubsection{Entity-type Information}
In biomedical literature, the acronym \textbf{ER} has multiple meanings: out of the $124$ mentions we found in $20$K recent PubMed abstracts,  $47$ refer to the gene ``estrogen receptor", $70$ refer to the organelle ``endoplasmic reticulum" and $4$ refer to the ``emergency room" in hospital. We use t-SNE \cite{maaten2008visualizing} to visualize different contextualized embeddings of these mentions in Figure \ref{fig:entity}. 

\noindent \textbf{In Domain v.s. General Domain:} For general ELMo, by far the strongest signal separating the mentions is whether they appear inside or outside parentheses.
This is not surprising given the recurrent nature of LSTM and language modeling training objective for learning these embeddings.
BioELMo does a better job of grouping mentions of the same entity (ER as estrogen receptor) together, which is clearly helpful for the NER task.

ER mentions of the same entities cluster better by BioBERT than general BERT: there are two major clusters corresponding to estrogen receptor and endoplasmic reticulum by BioBERT as indicated by the dashed circles, while entities of different types are scattered almost evenly by BERT.

\noindent \textbf{BioELMo v.s. BioBERT:} Clearly BioELMo better clusters entities from the same types together. Unlike ELMo/BioELMo, Whether the ER mention is inside parentheses doesn't affect BERT/BioBERT representations. It can be explained by encoder difference between ELMo and BERT: For ELMo, to predict `)' in forward LM, representations of token `ER' inside the parentheses need to encode parentheses information due to the recurrent nature of LSTM. For BERT, to predict `)' in masked LM, the masked token can attend to `(' without interacting with `ER' representations, so BERT `ER' embedding does't need to encode parentheses information.

\begin{table*}[htbp]
\centering
\small
\vspace*{6mm}
\begin{tabular}{lccccccc}
\toprule
\multirow{2}{*}{\textbf{Relation Type}} & \multicolumn{7}{c}{\textbf{NN w/ Representation of Same Type (\%)}} \\
\cmidrule(l){2-8} 
& \textbf{BioELMo} & \textbf{ELMo}  & \textbf{BioBERT-tog} & \textbf{BioBERT} & \textbf{BERT-tog} & \textbf{BERT} & \textbf{Biomed w2v}\\
\midrule
\textbf{disease-symptom} & \textbf{54.2} & 52.1 & 44.5 & 38.8 & 34.2 & 37.0 & 40.9\\
\textbf{disease-drug} & 32.8 & \textbf{34.4} & 26.1 & 17.9 & 27.7 & 22.6 & 23.6\\
\textbf{number-indication} & 70.5 & 63.9 & 47.0 & 45.3 & 48.1 & 49.5 & \textbf{74.4}\\
\textbf{synonyms} & \textbf{63.6} & 56.4 & 60.8 & 55.8 & 56.4 & 52.8 & 51.7\\
\midrule
\textbf{All} & \textbf{57.5} & 53.3 & 47.1 & 42.1 & 43.3 & 42.5 & 49.5\\
\midrule
\textbf{Subset Accuracy (\%)} & \textbf{73.9} & 62.8 & 71.4 & 65.0 & 65.8 & 64.5 & 69.7 \\
\bottomrule
\end{tabular}
\caption{Average proportion of nearest neighbor (NN) representations that belong to the same type for different embeddings, averaged over three random seeds. Biomed w2v performs best for number-indication relations, probably because it uses a vocabulary of over $5$M tokens, in which about $100$k are numbers. Subset accuracy denotes the probing task performance in the subset of MedNLI test set used for this analysis.}
\label{tab:relation}
\end{table*}

\subsubsection{Relational Information} 
We manually examine all test instances with the ``entailment" label in MedNLI, and found 78 token pairs across the premises and hypotheses which strongly suggest entailment. Among them, 22 are disease-symptom pairs, 13 are disease-drug pairs, 19 are numbers and their indications (e.g.: 150/93 and hypertension) and 24 are synonyms or closely related concepts (e.g.: Lasix\textsuperscript{\textregistered} and diuretic).
Figure~\ref{fig:mednli_relation} shows an example of disease-drug relationship. We hypothesize that a model is required to encode relation information to perform well in MedNLI task. We evaluate relation representations from different embeddings by nearest neighbor (NN) analysis: For each distributed relation representation (Eq.~\ref{eq:1}) of these token pairs, we calculated the proportions of its five nearest neighbors that belong to the same relation type. We report the average proportions in table \ref{tab:relation} and use it as a metric to measure the effectiveness of representing relations by different embedding schemes. We also show model performance for this subset (78 instances for relation analysis) in table \ref{tab:relation}. The trends of subset accuracy moderately correlate with the NN proportions (Pearson correlation coefficient $r=0.52$).

\noindent \textbf{In Domain v.s. General Domain:} For all relations, BioELMo is significantly\footnote{Significance is defined as $p<0.05$ in two-proportion z test.} better than ELMo  in representing same relations closer to each other, while there is no significant difference between BioBERT and BERT. This indicates that even pre-trained by in-domain corpus, as fixed feature extractor, BioBERT still cannot effectively encode biomedical relations compared to BERT.

\noindent \textbf{BioELMo v.s. BioBERT:} BioELMo significantly outperforms BioBERT and even BioBERT-tog for all relations. This explains why BioELMo does better than BioBERT in the probing task: BioELMo better represents vital biomedical relations between tokens in premises and hypotheses.


\section{Conclusion}
We have shown that BioELMo and BioBERT representations are highly effective on biomedical NER and NLI, and BioELMo works even without complicated downstream models and outperforms untuned BioBERT in our probing tasks. 
This effectiveness comes from its ability as a fixed feature extractor to encode entity types and especially their relations, and hence we believe they should benefit any task which requires such information. 

A long-term goal of NLP is to learn universal text representations. Our probing tasks can be used to test whether learnt representations effectively encode entity-type or relational information. Moreover, comprehensive characterizations of BioELMo and BioBERT as fixed feature extractors would also be an interesting further direction to explore.



\section{Acknowledgement}
We are grateful for the anonymous reviewers who gave us very insightful suggestions.
Bhuwan Dhingra is supported by a grant from Google.

\bibliography{naaclhlt2019}
\bibliographystyle{acl_natbib}

\end{document}